\documentclass{article}

\usepackage{amsmath,amsfonts,bm}

\def\eqref#1{equation~\ref{#1}}

\def\1{\bm{1}}

\def\mA{{\bm{A}}}
\def\mB{{\bm{B}}}

\def\mK{{\bm{K}}}

\def\mQ{{\bm{Q}}}

\def\mT{{\bm{T}}}

\def\mV{{\bm{V}}}
\def\mW{{\bm{W}}}

\DeclareMathAlphabet{\mathsfit}{\encodingdefault}{\sfdefault}{m}{sl}
\SetMathAlphabet{\mathsfit}{bold}{\encodingdefault}{\sfdefault}{bx}{n}

\def\sD{{\mathbb{D}}}

\def\sS{{\mathbb{S}}}

\def\sZ{{\mathbb{Z}}}

\def\emA{{A}}

\newcommand{\softmax}{\mathrm{softmax}}

\usepackage{microtype}
\usepackage{graphicx}
\usepackage{booktabs} 

\usepackage{hyperref}

\usepackage[accepted]{icml2024}

\usepackage{amssymb}
\usepackage{mathtools}

\usepackage[capitalize,noabbrev]{cleveref}

\usepackage{hyperref}
\usepackage{url}
\usepackage[T1]{fontenc}\usepackage{tikz}
\usepackage{amsmath, amsfonts}
\usetikzlibrary{arrows.meta}
\usepackage{float}
\usepackage{svg}
\usepackage{import}
\usepackage{pgfplots}
\pgfplotsset{width=7cm}
\usepackage{multirow}
\usepackage{makecell}
\usepackage{booktabs, multirow, tabularx}
\usepackage{caption}
\usepackage{subcaption}
\usepackage{pifont} 
\usepackage{pdfpages}
\usepackage{enumitem}

\usepackage{tablefootnote}

\usepackage{amsthm}

\usetikzlibrary{matrix}
\usetikzlibrary{positioning, matrix, arrows.meta}
\usetikzlibrary{calc}
\tikzset{square arrow/.style={to path={-- ++(0,.25) -| (\tikztotarget)}}}

\theoremstyle{plain}
\newtheorem{theorem}{Theorem}[section]

\theoremstyle{definition}
\newtheorem{definition}[theorem]{Definition}

\theoremstyle{remark}

\usepackage[textsize=tiny]{todonotes}

\icmltitlerunning{From Interpolation to Extrapolation: Complete Length Generalization for Arithmetic Transformers}

\begin{document}

\twocolumn[
\icmltitle{From Interpolation to Extrapolation: Complete Length Generalization for Arithmetic Transformers}




\begin{icmlauthorlist}
\icmlauthor{Shaoxiong Duan}{rdfz}
\icmlauthor{Yining Shi}{rdfz}
\icmlauthor{Wei Xu}{iiis}

\end{icmlauthorlist}

\icmlaffiliation{iiis}{Institute for Interdisciplinary Information Sciences, Tsinghua University}
\icmlaffiliation{rdfz}{RDFZ, Beijing, China}

\icmlcorrespondingauthor{Shaoxiong Duan}{shaoxiongduan@gmail.com}

\icmlkeywords{Machine Learning, ICML}

\vskip 0.3in
]



\printAffiliationsAndNotice{}  

\begin{abstract}

In this paper, we investigate the inherent capabilities of transformer models in learning arithmetic algorithms, such as addition and parity. Through experiments and attention analysis, we identify a number of crucial factors for achieving optimal length generalization. We show that transformer models are able to generalize to long lengths with the help of targeted attention biasing. In particular, our solution solves the Parity task, a well-known and theoretically proven failure mode for Transformers. We then introduce Attention Bias Calibration (ABC), a calibration stage that enables the model to automatically learn the proper attention biases, which we show to be connected to mechanisms in relative position encoding. We demonstrate that using ABC, the transformer model can achieve unprecedented \emph{near-perfect} length generalization on certain arithmetic tasks. In addition, we show that ABC bears remarkable similarities to RPE and LoRA, which may indicate the potential for applications to more complex tasks. \footnote{Code available at \url{https://github.com/shaoxiongduan/AttentionBiasCalibration}.}
\end{abstract}


\section{Introduction}

Large Language Models (LLMs) exhibit remarkable capabilities that offer promising insights into the development of Artificial General Intelligence (AGI). Having a world model, a representation or simulation of the external environment in which an intelligent agent operates, is considered essential for AGI. However, it is not clear whether, and to what extend, LLMs learn a world model, or just memorize ``surface statistics'' \citep{li2022emergent,bender-koller-2020-climbing}.

On the other hand, human acquire elements of our own world model, principles like the law of universal gravitation, through inductive learning: inferring general rules from finite number of examples or observations. \footnote{Note that we are talking about the process that a rule is discovered for the first time by human, not the process of an individual acquiring the knowledge, which may be through education.} It is known that inductive learning requires inductive biases \citep{Mitchell80}, additional assumptions that are independent of the data. This is because any finite number of training samples has infinite possible continuations corresponding to different generation rules. This constraint holds even for human learning. As David Hume stated, in his study of the problem of induction in \emph{A Treatise of Human Nature}, what we observe are nothing but sequences of ``constant conjunction'' and it is the human mind that imputes causation.  

Thus, we believe that, in order to understand and achieve AGI, it is important to understand how the model performs inductive learning and, more importantly, what are the effective ways to enforce inductive bias so that the model could discover the rule we desire (recall that there are infinite number of rules that could generate the training set). 

We conduct our study in the setting of the Transformer architecture and arithmetic tasks. Transformer has been the fundamental building block of many SOTA solutions across a wide range of machine learning tasks. And learning arithmetic algorithms presents a unique set of inductive learning tasks which can be seen as language transduction tasks, where the goal is to learn the underlying generation rules \citep{deletang2023neural}.The existence of such explicit generation rules provides a convenient setting where we can examine the internal mechanisms of the model.   

We use length generalization as an indicator to differentiate successful learning from memorization of  surface statistics. Length generalization, or \emph{extrapolation}, is defined as  ``a model’s ability to continue performing well as the number of input tokens during validation increases beyond the number of tokens on which the model was trained'' \citep{press2022train}. Many models achieve good accuracy with small inputs but fail to produce meaningful results with long inputs \citep{ruoss-etal-2023-randomized}. 
We define successful learning as complete length generalization, which is an indication that the model truly acquires the generation rules of the sequences. Specifically, 
\begin{definition}
\label{def.ext}
A model achieves complete length generation, or extrapolation, if it maintains at least 99\% accuracy when tested on samples with length at least 10 times the training length. 
\end{definition}
We set our investigation under the following goal and condition:
\begin{enumerate}
\item Complete generalization: The model must achieve complete generalization as defined by definition \ref{def.ext}. 
\item Learnable architecture: We only study mainstream architectures and methods (e.g., common PEs) that are learnable using regular optimization algorithms (e.g., SGD).
\end{enumerate}

The condition is consistent with the most powerful models. Surprisingly, within such a setting, ``simple'' arithmetic tasks such as addition are actually very hard for Transformers. We elaborate the difficulties in section \ref{sec.setting}.

In this work, we draw attention to the stage of model \emph{interpolation}. Interpolation is a special form of in-distribution generalization, defined as a model's ability to perform well on examples that are novel but with lengths within the same range as those samples from the training set. We show in section \ref{sec.abc} that the patterns that the model acquires during the interpolation stage can be used as a form of inductive biases to re-train the model to achieve \emph{extrapolation}.



The contributions of this work include:
\begin{itemize}

\item We show that attention biasing is an effective way to enforce inductive bias for Transformer architecture.

\item We are the \emph{first} Transformer-based architecture to obtain complete generalization on a number of arithmetic tasks: successor function, parity, addition, and a restricted version of multiplication. Our models produce results with 100\% accuracy up to 60 digits.


\item We show that, for the tasks that we study, (the right) attention is indeed all you need. Transformer can perform well as long as it attends to the right tokens. And we identify a few key factors in achieving such proper attention. Among them, we introduce Cyclic Position Indexing (CPI), a new position index scheme that allows tasks relying on localized attention such as addition to generalize.

\item Based on our findings, we introduce \emph{attention bias calibration} (ABC), a process that automatically collects attention patterns learned from training data and extends them to long lengths. We show that this automizes the above mechanisms. In addition to that, we show ABC's relation to RPE and LoRA, which indicates the potential for its applications to more complicated tasks.

\end{itemize}

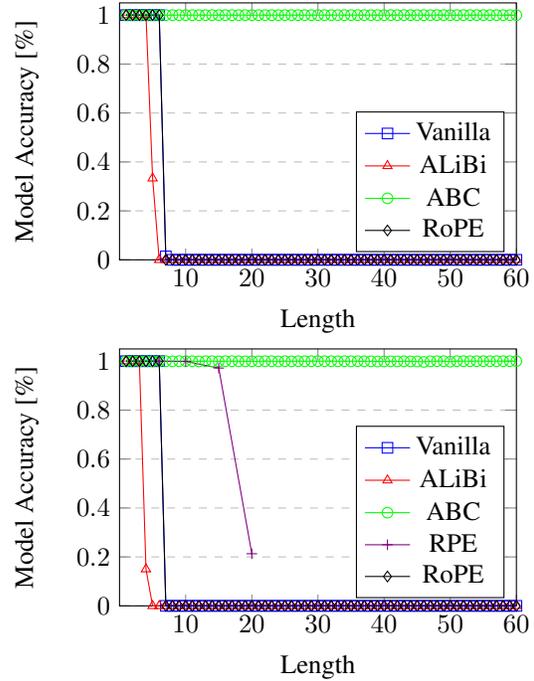
\begin{figure}
\centering

\begin{subfigure}[b]{0.4\textwidth}
\centering
\begin{tikzpicture}[scale=1]
\begin{axis}[
    xlabel={Length},
    ylabel={Model Accuracy [\%]},
    xmin=0, xmax=60,
    ymin=0, ymax=1.05,
    xtick={10, 20, 30, 40, 50, 60},
    ytick={0,0.2,0.4,0.6,0.8,1.0},
    legend pos= south east,
    ymajorgrids=true,
    grid style=dashed,
width=\textwidth,height=5cm]
\addplot[color=blue,mark=square] table[y= acc_vanilla]{cnt_extrapolation.dat}; \addlegendentry{Vanilla}
\addplot[color=red, mark=triangle] table[y=acc_alibi] {cnt_extrapolation.dat}; \addlegendentry{ALiBi}
\addplot[color= green, mark=o] table[y=acc_ex] {cnt_extrapolation.dat}; \addlegendentry{ABC}
\addplot[color=black, mark=diamond] table[y=acc_rope] {cnt_extrapolation.dat}; \addlegendentry{RoPE}
\end{axis}
\end{tikzpicture}
\end{subfigure}%
\hfill
\begin{subfigure}[b]{0.4\textwidth}

\begin{tikzpicture}[scale=1]
\begin{axis}[
    xlabel={Length},
    xmin=0, xmax=60,
    ymin=0, ymax=1.05,
    xtick={10, 20, 30, 40, 50, 60},
    ytick={0,0.2,0.4,0.6,0.8,1.0},
    ylabel={Model Accuracy [\%]},
    legend pos=south east,
    ymajorgrids=true,
    grid style=dashed,
width=\textwidth,height=5cm]
\addplot[color=blue,mark=square] table[y= acc_vanilla]{add_extrapolation.dat}; \addlegendentry{Vanilla}
\addplot[color=red, mark=triangle] table[y=acc_alibi] {add_extrapolation.dat}; \addlegendentry{ALiBi}
\addplot[color= green, mark=o] table[y=acc_ex] {add_extrapolation.dat}; \addlegendentry{ABC}
\addplot[color= violet, mark=+] coordinates {(6,1)(10,0.999)(15,0.972)(20,0.213)}; \addlegendentry{RPE}
\addplot[color=black, mark=diamond] table[y=acc_rope] {cnt_extrapolation.dat}; \addlegendentry{RoPE}
\end{axis}
\end{tikzpicture}

\end{subfigure}%
\caption{Extrapolation results for models trained on $L_\mathit{int} \leq 6$ on \textbf{\texttt{Successor}} (top) and \textbf{\texttt{Addition}} (bottom). Length is measured in the number of digits of one operand.}
\label{fig.main}
\end{figure}

Figure \ref{fig.main} summarizes the generalization that our ABC scheme achieves on two of the tasks, with comparisons against popular alternatives such as ALiBi, RoPE, etc. ABC is the only solution achieving perfect generalization. We obtain similar results on other tasks which will be discussed in detail in section \ref{sec.main_results}.

\section{Related Work}
Length generalization for Transformers is a very hot topic in other areas. And indeed we draw on many of their inspirations. In this section we briefly summarize some of the most influential works. In section \ref{sec.setting} we will further elaborate on the line of work on arithmetic algorithms learning to better situate our work.

Existing works on Transformer length generalization have been mainly focusing on two aspects: positional encoding (PE) or/and attention bias (AB).

\textbf{Relative Position Encoding}.
Relative position encoding (RPE) relies on the relative distance between tokens to construct position embeddings. This approach is first proposed by \citet{shaw-etal-2018-self} and has shown to produce significant improvements over absolute positional encoding in machine translation tasks \citep{shaw-etal-2018-self}. This leads to its application in numerous machine learning models and the development of multiple variations such as Transformer-XL \citep{dai2019transformerxl} and RoPE \citep{su2022roformer}.

\textbf{Attention Biasing}.
Attention biasing, on the other hand, adds a bias directly to the attention matrix, allowing the model to extrapolate to longer lengths efficiently. First introduced as ALiBi (Attention with Linear Biases) by \citet{press2022train}, it is quickly followed by similar models such as KERPLE \citep{chi2022kerple}, and Sandwich \citep{chi-etal-2023-dissecting}, all showing certain improvement in length extrapolation. Other forms of biases include sliding window \citep{beltagy2020longformer} and its variations. Compared to other relative position encoding schemes, attention biasing typically demands less computational resources.

These two lines of work are closely related and there are extensive studies on their effectiveness. \footnote{Please see \citet{10.1162/coli_a_00445} for a comprehensive review of methods to incorporate position information into Transformer models.} However,  the results are mixed. On one hand, the popular belief is that relative PEs \citep{shaw-etal-2018-self,dai2019transformerxl,su2022roformer} are more effective in length generalization than absolute variants \citep{attentionisallyouneed}. On the other hand, however, some works (e.g., \citet{kazemnejad2023impact}) point out that such a conclusion is obtained by using language modeling perplexity as the sole metric, which may not reflect actual performances on downstream tasks. In fact, \citet{kazemnejad2023impact} show that, on a collection of reasoning and mathematical tasks, No Positional Encoding (NoPE) actually performs the best. Likewise, \citet{deletang2023neural} show that state-of-the-art PE or AB methods do not help Transformer extrapolate on arithmetic tasks.
  



\section{Setup}
\label{sec.setup}

\subsection{Tasks}
\label{sec.Tasks}
Let $\mathbb{N} = \{0, 1, 2, \ldots \}$ be the set of natural numbers. We consider the following arithmetic tasks:
\begin{itemize}
\item \textbf{Successor function}: Maps a natural number to the next one: $S(n) = n + 1$ for $n \in \mathbb{N}$.
\item \textbf{Addition}: $y = x_1 + x_2$ for $x_1, x_2 \in \mathbb{N}$.
\item \textbf{Parity}: Given $x \in \mathbb{N}$, this operation returns 1 if the binary representation of $x$ contains an odd number of 1's, and 0 otherwise.
\item \textbf{$N \times 1$}: $y = x_1 \times x_2$ for $x_1 \in \mathbb{N}$ and $ x_2 \in \{0, 1, \ldots, 9\}$. This is a restricted form of multiplication where one of the operands is restricted to single-digit.
\end{itemize}
These tasks are well-known examples in the theory of computation. The seemingly trivial \textbf{\texttt{Successor}} function is the basic component of Peano axioms, which formalize the structure of the natural numbers. Using the digit-based representation, \textbf{\texttt{Successor}}, \textbf{\texttt{Addition}} and \textbf{\texttt{$N \times 1$}} all  belong to the Type-1  context-sensitive (CS) category of the Chomsky hierarchy. \textbf{\texttt{Parity}}, on the other hand, is in Type-3 Regular (R) category \citep{deletang2023neural}, since it can be solved with a 2-state finite-state machine. \textbf{$N \times 1$} is a task that is specifically constructed to test if the methods we develop could be extended to more complex operations such as multiplication. Unlike addition, carry is a little more complex in multiplication and involves multiple values (i.e., carry can be any digit among $\{0, 1, \ldots, 9\}$). Restricting one operand to be single-digit results in a simpler attention pattern that allows for easy analysis. 

 
\subsection{Difficulties}
 \label{sec.setting}
 
There are specially constructed architectures such as \citet{chiang-cholak-2022-overcoming,Deshpande2021RecTAR} that achieve generalization on some of the tasks we study. Clearly they do not conform to our setting. To the best of our knowledge, there is no previous work achieving true generalization under  our setting. Very recent works of \citet{ruoss-etal-2023-randomized} and \citet{deletang2023neural} conduct extensive empirical studies (6,000 and 20,910 models, respectively, across 15  tasks). In particular, \citet{deletang2023neural} consider five major positional encodings: none, classical sin/cos \citep{attentionisallyouneed}, RoPE \citep{su2022roformer}, ALiBi \citet{press2022train}, and Transformer-XL \citep{dai2019transformerxl}, and report the best performing configuration.  Both \citet{deletang2023neural} and \citet{ruoss-etal-2023-randomized} obtain slightly better than random accuracy on \textbf{\texttt{Parity}}, and 54.3\% and 64.5\% on binary addition, respectively. Neither generalizes. \citet{zhou2023algorithms} try to generalize by increasing training length and achieve good performance up to 10 additional digits, which does not conform to our standard of generalization. \citet{yang2023gpt} enhance LLM 's arithmetic capability by fine-tuning pre-trained language models. Their solution does not extrapolate, since their evaluation dataset is generated from the same distribution as the training dataset.

\textbf{The Theoretical Impossibility.}  In fact, \citet{hahn-2020-theoretical} shows that, for self-attention models with soft attention (e.g., a typical Transformer), the change in activation at the decoder layer that changing a single input symbol can cause is bounded by $O(1/n)$ where $n$ is the input length (Lemma 5). This means that for tasks that are sensitive to changes of a small number of input symbols, such as \textbf{\texttt{Parity}}, Transformer models cannot make accurate predictions for long sequences, as softmax is unable to produce very different predictions for inputs that result in similar activations. 

\textbf{\texttt{Parity}} is the simplest non-counter-free regular language, the lowest layer of the Chomsky hierarchy. This limitation may imply an impossibility for Transformer to solve any regular language that is not counter-free \citep{hahn-2020-theoretical}. \citet{chiang-cholak-2022-overcoming} prove that there exists a specially crafted Transformer construction that can achieve perfect \textbf{\texttt{Parity}}, but such a construction is not learnable.

To overcome the limitation, architectural changes or other means are inevitable. Ours is the \emph{first} learnable Transformer that obtains perfect accuracy and length generalization for the \textbf{\texttt{Parity}} task. We overcome the difficulties of \citet{hahn-2020-theoretical} by applying the ``scaffolding''  methods introduced later. Mathematically, our methods can be seen as allowing the model to produce (partial) results based on only local tokens, effectively reducing $n$ to a constant window size. 

\subsection{Tokenization and Problem Representation}
\label{subsec.tok}


Unlike some previous works that study the arithmetic operations within a finite group and treat each number as a token (e.g., \citet{power2022grokking}), we use a character-based tokenizer which allows us to represent an infinite number of integers using a finite set of tokens, enabling unbounded extrapolation testing. Thus in our study the numbers and operators are all encoded in their natural form and we use a  vocabulary of 15 tokens: \{\texttt{0}, ..., \texttt{9}, \texttt{+}, \texttt{*}, \texttt{\$}, \texttt{\&}, \texttt{@}\}, where the last three characters represent \texttt{SOS, EOS}, and \texttt{PAD}, respectively.

With our tokenization, all tasks are naturally sequence-to-sequence except for \textbf{\texttt{Parity}} which is classification. We turn \textbf{\texttt{Parity}} into a sequence-to-sequence task as follows. Let $x_n\ldots x_1$ be the input sequence of a binary string where $x_i \in \{0, 1\}$, the target is generated as $y_1 = x_1, y_i = y_{i-1} \otimes x_{i}$ for $i = 2, \ldots, n$, where $\otimes$ denote bitwise xor. This can be seen as a form of scratch pad which may appear to simplify the problem. However, it only provides explicit information about the correct computation steps but the task is actually harder due to the compounding-of-errors effect. We provide analysis and empirical results in section \ref{sec.abs_results}. 

To facilitate learning, we pad each operand with 0's to the left to a fixed length. In addition to that, we reverse the ordering of the tokens in the output sequence to match the natural generation process. For example, $0123+0748 \rightarrow 1780$.

\subsection{Model Configuration}
Since our tasks are sequence-to-sequence, we choose an encoder-decoder architecture, with 1 encoder layer and 6 decoder layers, all with 8 attention heads. The embedding size is 128 and feed forward size 512. We tried models with a number of different sizes and found no significant difference across all variations that could converge. We settled on the model above and did not pursue the configuration with the optimal size. 

We train our models using cross-entropy loss and Adam optimizer, with learning rate	$10^{-5}$ and a dropout of $0.3$. For training for interpolation, we generate a random permutation $\Pi$ of numbers in the range [0, $2^{20}$] and split the set by a 7:1 ratio for training and validation. For binary operations such as \textbf{\texttt{Addition}}, both operands are drawn independently from $\Pi$. Thus both the training and validation data sets consist of mainly 6-digit numbers, in base 10, with less than 5\% 7-digit numbers. We denote $L_\mathit{int}$ the length of input, measured as the maximum number of digits in the operand(s), during the interpolation phase. Note that length refers to the number of digits in the operands, not the total input sequence length. 
 
For extrapolation testing, for each length $L$,  we randomly sample $\min(10^L-10^{L-1}, 10000)$ numbers of length $L$ and compute the accuracy on these samples. For \textbf{\texttt{Parity}} which deals with binary sequences, we still generate train and test numbers in the way described above and convert them into binary sequences for training and testing. Since a number's length in decimal is proportional to its length in binary, the 10x length expansion is preserved in either base. The model's output is considered accurate if and only if it exactly matches the correct label sequence. 

We use greedy decoding for all inferences.


\section{(The Right) Attention is All You Need}
\label{sec.right_attn}

To develop our ideas, we first train vanilla Transformers with some commonly used length generalization methods, including the original sinusoidal positional encoding, ALiBi, and RoPE. The results on \textbf{\texttt{Successor}} and \textbf{\texttt{Addition}}, together with the performance of our ABC scheme,  have been shown in  figure \ref{fig.main} at the beginning of the paper.  All models achieve some levels of interpolation but none could extrapolate beyond training length. RoPE and vanilla Transformer perform almost identically, dropping precipitously to almost 0 accuracy once the length goes beyond 6. We observe similar patterns with other tasks.

\begin{figure*}
\begin{subfigure}{.16\textwidth}
\centering
\includegraphics[height=1.8cm]{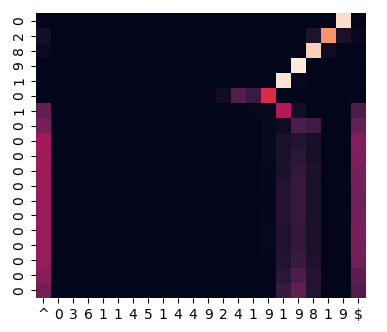} %
\caption{Cross Attn}
\end{subfigure}
\begin{subfigure}{.16\textwidth}
\centering
\includegraphics[height=1.8cm]{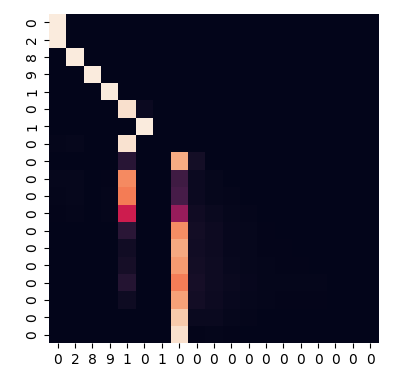} %
\caption{Self Attn}
\end{subfigure}
\vrule
\begin{subfigure}{.33\textwidth}
\centering
\includegraphics[height=1.8cm]{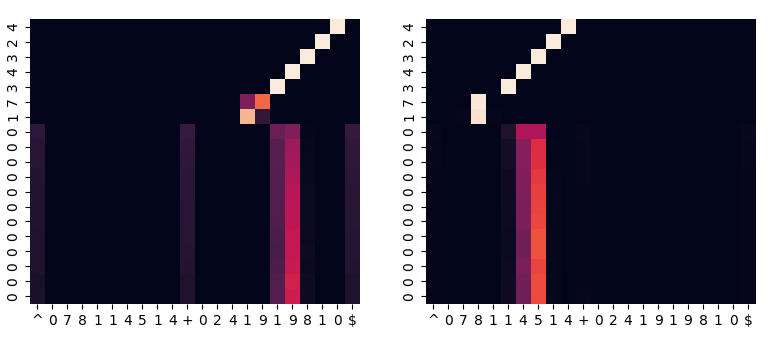} %
\caption{Decoder Cross Attention}
\end{subfigure}
\begin{subfigure}{.33\textwidth}
\centering
\includegraphics[height=1.8cm]{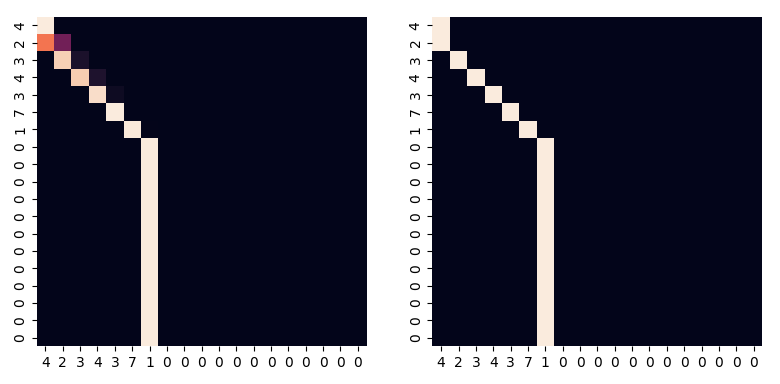} %
\caption{Decoder Self Attention}
\end{subfigure}
\caption{Attention heat maps for  \textbf{\texttt{Successor}} (Left) and \textbf{\texttt{Addition}} (Right).}
\label{fig.attn}
\end{figure*}

To figure out the causes of failure to extrapolate, we extract and analyze the attention weights of the vanilla model on \textbf{\texttt{Successor}} and \textbf{\texttt{Addition}}. Figure \ref{fig.attn} shows the attention heat maps of one specific head in the last decoder layer when decoding \textbf{\texttt{Successor}} and \textbf{\texttt{Addition}} tasks. Lighter colors represent higher weights. More detailed analysis is presented in appendix \ref{sec.attn} but the patterns are very clear:  the vanilla Transformer correctly learns the right attention patterns up to the training length and fails beyond that. This correlates perfectly with the extrapolation performance shown in figure \ref{fig.main}.  

\subsection{Attention Bias Scaffolding}
\label{sec.abs}

 
In this section, we introduce a number of methods that guide the model to attend to the right places. Assisting model learning is a common practice. Relevant techniques include inputs combination \citep{Libovick2018inputcombo}, ``arithmetic prompting'' \citep{zhou2022teaching}, representation transformation \citep{nogueira2021investigating}, scratch pad \citep{lee2023teaching}, etc. Indeed, most of our methods are drawn from these toolboxes as well. However, we use them to target directly at the attention, thus we call our approach Attention Bias Scaffolding (ABS). In the following we briefly summarize the two most effective ones. A detailed treatment including visualizations can be found in appendix \ref{sec.abs_details}.

\textbf{Windowed Attention Biasing}. The idea was developed by Longformer \citep{beltagy2020longformer}. The basic intuition is that the most important local dependency is typically restricted by a limited range.  In the context of arithmetic algorithm learning, often local context is the \emph{sole} determining factor, \footnote{This is why the arithmetic calculation algorithms that we humans use could be applied to arbitrarily long inputs.} which  can be captured by a sliding window of width $w$ \citep{beltagy2020longformer}.  

\textbf{Cyclic Position Indexing (CPI)}. Position indexing refers to how we identify each individual position. The simplest way is just to index them $0, 1, \ldots$.  As our tasks have very restricted dependency contexts which are localized by the windowed attention biasing, the model only needs a way to differentiate positions within the context window thus long position indexing is not necessary. And our empirical study shows that it can be harmful sometimes. Therefore we propose Cyclic Position Indexing (CPI): Let $i$ be the position index of a token. We select a period parameter $T$ and convert token positions to $i \mod T$ before entering into the model. Essentially indices are cycling around back when they get large. 


\subsection{Results of ABS}

To evaluate the effectiveness of the above mechanisms, we conduct extensive experiments on each of the arithmetic tasks with a number of their combinations. Results and detailed discussion are presented in table \ref{tab.abs} in appendix \ref{sec.abs_results}. We summarize the main findings here:

\begin{itemize}
\item None of the previous works achieves extrapolation on any of the tasks. 
\item Our solutions (windowed attention biasing + CPI) achieve complete length generalization on all tasks, maintaining 100\% accuracy up to 60 digits. 
\item Unary tasks (\textbf{\texttt{Successor}} and \textbf{\texttt{Parity}}) appear to be not relying on any positional embedding at all once the windowed attention biasing is in place.
\item For binary tasks (\textbf{\texttt{Addition}} and \textbf{\texttt{$N \times 1$}}), there appears to be some bad interaction between the original sinusoidal PE and windowed attention biasing. Their combination achieves only interpolation but not extrapolation. 
\end{itemize}



The above suggests that the right attention is the key to achieving good generalization (thus the title of this section). 

\section{Attention Bias Calibration (ABC)} 
\label{sec.abc}

Having demonstrated the important role of correct attention in the transformer model's learning, we introduce Attention Bias Calibration (ABC), an automatic process that extends the working attention patterns of a model that achieves successful interpolation to arbitrary lengths while preserving its near-perfect performance. The idea is, a model trained to full interpolation must be able to produce the right attention pattern on interpolation data (see section \ref{sec.attn}), which captures the local dependencies for recurrent arithmetic algorithms. ABC extracts and aggregates the attention weights and uses them as attention bias, like \citet{press2022train}, to fine-tune the model for long inputs. Similar to the scaffolding in section \ref{sec.abs}, ABC is also a kind of inductive bias, but it is fully automatic.


 

Let $m \times n$ be the dimensions of the attention matrix of a model that has interpolated and $M \times N$ the dimensions that we would like to extrapolate to. It should hold that $m < M$ and $n < N$.  ABC proceeds in the following steps:

\textbf{1. Training for Interpolation:} First we train a vanilla transformer model $\displaystyle \mT_{int}$  on the dataset $\displaystyle \sS_{int}$ until it is capable of interpolation. By this point, the accuracy of $\displaystyle \mT_{int}$ should be near perfect. Then we use $\displaystyle \mT_{int}$ to decode a random subset of training samples $\displaystyle \sS_{gen} \subset_R \sS_{int}$ and extract the attention weights. Because this process is identical for all heads, to simplify notation, we omit their indices. Let $x_k[i]$ be the embedding vector for the $i$-th token in sample $k$, the attention matrix is extracted as

\begin{displaymath}
A^{k}_{i,j}=x_k[i]\mW^{Q} {\mW^{K}}^{\top} {x_k[j]}^\top
\end{displaymath}

where $\mW^Q, \mW^K$  are parameter matrices in the last decoder layer of model $\displaystyle \mT_{int}$. 

\textbf{2. Attention Biases Computation:} We then  average the attention weights for all data in $\displaystyle \sS_{gen}$:
\begin{displaymath}
\bar{\mA} = \frac{1}{|\sS_{gen}|} \sum_{k=1}^{|\sS_{gen}|}  \mA^k 
\end{displaymath}

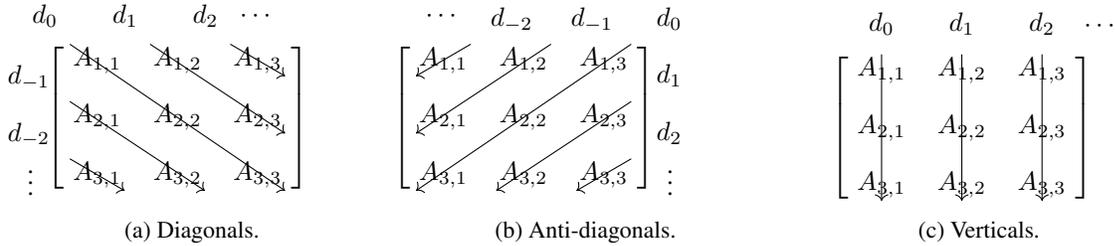
\begin{figure*}
\centering
\begin{subfigure}{.3\textwidth}
\begin{tikzpicture}
\matrix[matrix of math nodes,inner sep=1pt,row sep=1em,column sep=1em, left delimiter={[},right delimiter={]}] (M)
{
    {\displaystyle \emA}_{1,1} &  {\displaystyle \emA}_{1,2}&  {\displaystyle \emA}_{1,3}  \\
     {\displaystyle \emA}_{2,1} &  {\displaystyle \emA}_{2,2} &  {\displaystyle \emA}_{2,3} \\
     {\displaystyle \emA}_{3,1} &  {\displaystyle \emA}_{3,2}&  {\displaystyle \emA}_{3,3}  \\
}
;
\draw[->] (M-1-1.north west) -- (M-3-3.south east);
\draw[->] (M-1-2.north west) -- (M-2-3.south east);
\draw[->] (M-1-3.north west) -- (M-1-3.south east);
\draw[->] (M-2-1.north west) -- (M-3-2.south east);
\draw[->] (M-3-1.north west) -- (M-3-1.south east);
\node[above left=3pt and 1pt of M-1-1]{$d_0$};
\node[above left=3pt and 1pt of M-1-2]{$d_1$};
\node[above left=3pt and 1pt of M-1-3]{$d_2$};
\node[above=5pt of M-1-3]{$\cdots$};
\node[above left=1pt and 4pt of M-2-1]{$d_{-1}$};
\node[above left=1pt and 4pt of M-3-1]{$d_{-2}$};
\node[left=10pt of M-3-1]{$\vdots$};
\end{tikzpicture}
\caption{Diagonals.}
\label{fig.diag1}
\end{subfigure}
\begin{subfigure}{.3\textwidth}
\begin{tikzpicture}
\matrix[matrix of math nodes,inner sep=1pt,row sep=1em,column sep=1em, left delimiter={[},right delimiter={]}] (M)
{
    {\displaystyle \emA}_{1,1} &  {\displaystyle \emA}_{1,2}&  {\displaystyle \emA}_{1,3}  \\
     {\displaystyle \emA}_{2,1} &  {\displaystyle \emA}_{2,2} &  {\displaystyle \emA}_{2,3} \\
     {\displaystyle \emA}_{3,1} &  {\displaystyle \emA}_{3,2}&  {\displaystyle \emA}_{3,3}  \\
}
;
\draw[->] (M-1-3.north east) -- (M-3-1.south west);
\draw[->] (M-1-2.north east) -- (M-2-1.south west);
\draw[->] (M-1-1.north east) -- (M-1-1.south west);
\draw[->] (M-2-3.north east) -- (M-3-2.south west);
\draw[->] (M-3-3.north east) -- (M-3-3.south west);
\node[above right=2pt and 6pt of M-1-3]{$d_0$};
\node[above right=2pt and 6pt of M-2-3]{$d_1$};
\node[above right=2pt and 6pt of M-3-3]{$d_2$};
\node[above=5pt of M-1-1]{$\cdots$};
\node[above right=1pt and 4pt of M-1-2]{$d_{-1}$};
\node[above right=1pt and 4pt of M-1-1]{$d_{-2}$};
\node[right=10pt of M-3-3]{$\vdots$};
\end{tikzpicture}
\caption{Anti-diagonals.}
\label{fig.antidiag1}
\end{subfigure}
\begin{subfigure}{.3\textwidth}
\centering
\begin{tikzpicture}
\matrix[matrix of math nodes,inner sep=1pt,row sep=1em,column sep=1em, left delimiter={[},right delimiter={]}] (M)
{
    {\displaystyle \emA}_{1,1} &  {\displaystyle \emA}_{1,2}&  {\displaystyle \emA}_{1,3}  \\
     {\displaystyle \emA}_{2,1} &  {\displaystyle \emA}_{2,2} &  {\displaystyle \emA}_{2,3} \\
     {\displaystyle \emA}_{3,1} &  {\displaystyle \emA}_{3,2}&  {\displaystyle \emA}_{3,3}  \\
}
;
\draw[->] (M-1-1.north) -- (M-3-1.south);
\draw[->] (M-1-2.north) -- (M-3-2.south);
\draw[->] (M-1-3.north) -- (M-3-3.south);
\node[above=4pt of M-1-1]{$d_0$};
\node[above=4pt of M-1-2]{$d_1$};
\node[above=4pt of M-1-3]{$d_2$};
\node[above right=5pt and 2pt of M-1-3]{$\cdots$};
\end{tikzpicture}
\caption{Verticals.}
\label{fig.diag3}
\end{subfigure}
\caption{Examples of the different directions ABC explores.}
\label{fig.diag}
\end{figure*}

The next steps average attention weights along a number of lines within the elements of the matrix and extend them along those particular directions. We observe that attention patterns manifest themselves along lines of the attention matrix and these are the directions we expand them. Theoretically, we could explore any direction but empirically we find it suffices to only try the diagonal, the anti-diagonal, and the vertical lines. Figure \ref{fig.diag} visualizes the said directions, with line sums annotated on the sides.

For all directions we consider, let $l$ be the set of elements on a line, we perform the following steps:

\textbf{2.1. Averaging across Lines:} 
\begin{displaymath}
d_{l} = \frac{1}{|l|}\sum_{(i,j) \in l} \bar{A}_{i,j}
\end{displaymath}
This step effectively ``summarizes'' each line into a single value.

\textbf{2.2. Bias Matrix Extension:} Next we extend $\bar{\mA}$ into any arbitrary size $\tilde{\mA} \in \mathbb{R}^{M \times N}$ via:
\begin{equation}
\label{eq:3}
\tilde{A}_{i,j} =
\begin{cases}
	 \mathit{dropoff} (d_{l} - d_{max}),& \text{if } l \text{ exists in }\bar{\mA} \\
	-\inf, & \text{otherwise}
\end{cases}
\end{equation}

where $d_{max}$ is the maximum value of $d_l$'s among all the lines of $\bar{\mA}$, and 

\begin{displaymath}
 \displaystyle \mathit{dropoff}(x) = 
 \begin{cases}
	x,& \text{if } {x>\mathit{threshold}} \\
	-\inf, & \text{otherwise}
\end{cases}
\end{displaymath}

What this process does is actually very simple: for the elements along the extension of existing lines of $\bar{\mA}$, it first subtracts $d_{max}$  from $d_l$, then cuts off at a $threshold$. Elements not on the extensions of $\bar{\mA}$'s lines will be set to $-\inf$. For our task, the dropout threshold is set to $\kappa\sigma + \mu$, where $\sigma$ and $\mu$ are the standard deviation and the mean of all the $d_l$'s, respectively, and $\kappa$ is an empirically determined factor. We set $\kappa$ = 4.5 and 0.87 for cross and self attention, respectively. This results in very strict thresholds, meaning that it only preserves really strong patterns. For other tasks where patterns are not that obvious, a softer threshold value or even no dropout may be used.

\textbf{2.4. Finalization:} The final bias matrix $\tilde{\mA}$ is obtained by performing an element-wise $\max$ operation among the matrices from equation \ref{eq:3} across all directions. We then repeat for each of the heads, equipping them with independent biases.
If the final bias matrix consists of only $-\inf$'s, meaning that no pattern is picked up, we replace every $-\inf$ with $0$, effectively leaving it ``transparent''.

The complete and detailed algorithm is presented in appendix \ref{sec.alg}.

\textbf{3. Re-training with Attention Biases:} After the attention biases for each head have been constructed, we train another model on the same input sequences $\displaystyle \sS_{int}$ with the constructed attention biases added to the attention weights.
\begin{equation}
\label{eqn.abc}
\displaystyle A_{i,j}={x_i\displaystyle \mW^{Q}{\displaystyle \mW^{K}}^\top x_j^\top + \tilde{A}_{i,j}}, \quad  \tilde{\mA} \in \mathbb{R}^{M \times N}
\end{equation}
Note that in this work the bias matrices are obtained from the last decoder layer and applied to all layers during re-train. More flexible configurations such as per-layer bias could work better for more complex tasks.





\section{Main Results}
\label{sec.main_results}

A prerequisite of ABC is that the vanilla Transformer must be able to train to interpolate. Among the tasks we study, as discussed in section \ref{sec.setting}, \textbf{\texttt{Parity}} is apparently a failure. Thus we implement the vanilla Transformer (with sinusoidal PE), ALiBi, RoPE, and ABC,  and test on the rest of the tasks. Note that we do not use the input alignment method developed for ABS in section \ref{sec.abs_details}. The inputs to the model are in their ``natural'' form such as $0123+0748 \rightarrow 1780$.


The accuracy vs. input length curves of different models on \textbf{\texttt{Successor}} and \textbf{\texttt{Addition}} have been plotted in figure \ref{fig.main} at the beginning of this paper.  The overall performance on all tasks is summarized in table \ref{results-table}. We observe that ABC performs vastly superior to other models across all tasks, achieving near-perfect accuracies up to 60 digits. 

\begin{figure}[h]%
\centering
\includegraphics[scale=1,width=.5\textwidth]{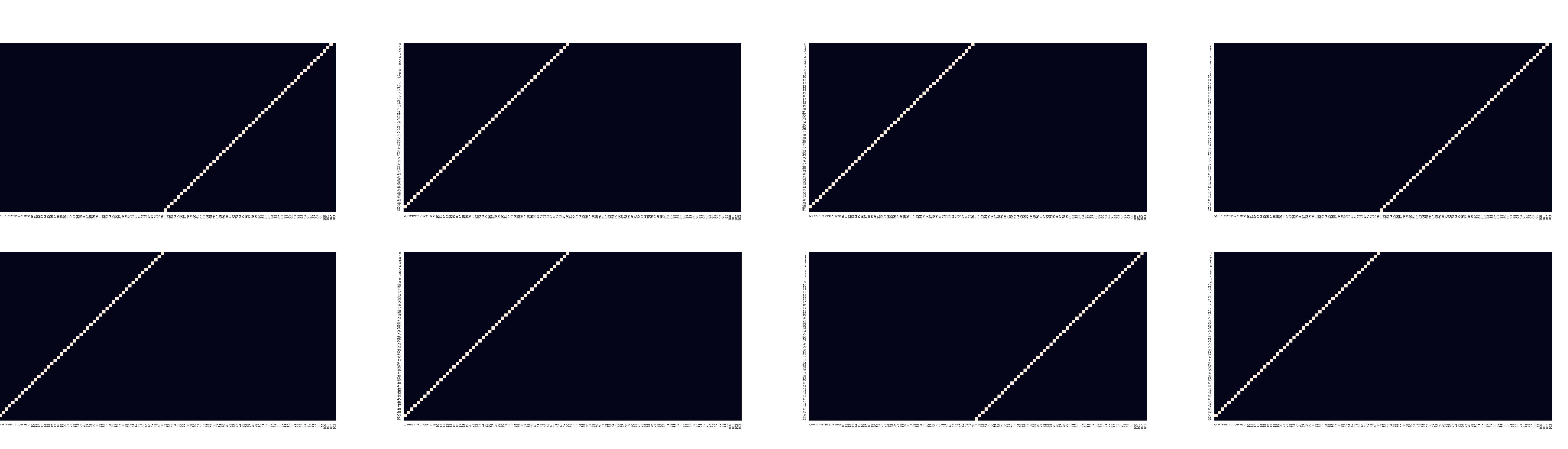}
\caption{ABC cross attention bias for \textbf{\texttt{Addition}}}
\label{fig:ABCbiasaddcross}
\end{figure}

\begin{figure}[h]%
\centering
\includegraphics[scale=.8,width=.5\textwidth]{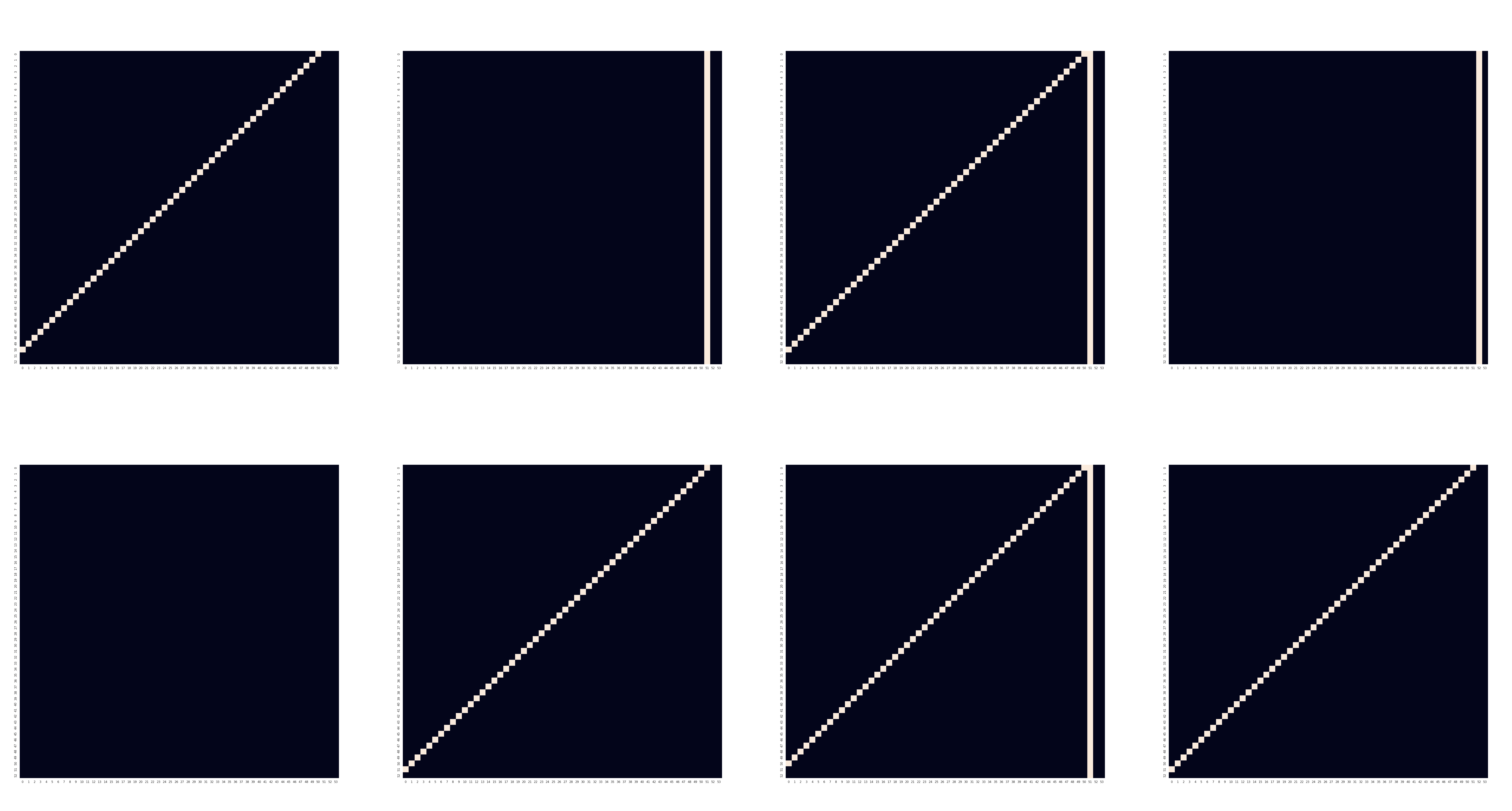}
\caption{ABC cross attention bias for \textbf{\texttt{$N \times 1$}}}
\label{fig:ABCbiasmultcross}
\end{figure}

Figure \ref{fig:ABCbiasaddcross} and \ref{fig:ABCbiasmultcross} visualize the cross attention bias matrices, one for each head, learned by ABC, for  \textbf{\texttt{Addition}} and \textbf{\texttt{$N \times 1$}}, respectively.  Since the most meaningful attention activities happen in cross-attention, where the model is attending to the input sequence, we do not show self-attention biases. Each color map is plotted using a colorbar scaling of [$\min_h (\displaystyle \tilde{\mA}_h)$, $\max_h (\displaystyle \tilde{\mA}_h)$] for each individual head. Head bias with a small variance will result in a ``transparent'' bias matrix with all 0's after drop-off, in which case the 0's are painted black.  Note that addition is a binary operation so the input length is twice the output sequence thus the matrices in figure \ref{fig:ABCbiasaddcross} are rectangles instead of squares.

\begin{table}[h]
\caption{Extrapolation results measured as percent accuracy (\%).  Numbers in bold show the best accuracies achieved for the corresponding input length limit.}
\label{results-table}
\begin{center}

\begin{tabular}{c r c c c c}
\toprule
\multicolumn{2}{c}{\bf}  &\multicolumn{4}{c}{\bf Length (Number of Digits)} \\ 
\multicolumn{1}{c}{\bf Task} &\multicolumn{1}{c}{\bf Model}  &\multicolumn{1}{c}{\bf 6}  &\multicolumn{1}{c}{\bf 10}  &\multicolumn{1}{c}{\bf 20} &\multicolumn{1}{c}{\bf 60} \\
\midrule
    		&Vanilla & \bf 100.0 & 0.0  &0.0 &0.0 \\
Successor   	&ALiBi &$1.3$ &0.0  &0.0 &0.0\\
           	&RoPE &\bf100.0 &0.0  &0.0 &0.0 \\
		&ABC &\bf 100.0 &\bf 100.0  &\bf100.0 &\bf100.0\\
\midrule
    		&Vanilla & \bf 100.0 & 0.0  &0.0 &0.0 \\
	  	&ALiBi &0.0 &0.0 &0.0  &0.0\\
Addition   	&RoPE &\bf 100.0 &0.0  &0.0 &0.0 \\
		&RPE$^*$ &\bf 100.0&$99.9$  &$21.3$ &N/A \\
		&ABC &\bf 100.0 &\bf 100.0  &\bf 99.9 & \bf 99.8\\
\midrule
   		&Vanilla & \bf 100.0 & 0.0 &0.0 &0.0 \\
$N \times 1$		 &RoPE &\bf100.0 &0.0 &0.0 &0.0 \\
		&ABC &\bf 100.0 &\bf 100.0  &\bf100.0 &\bf100.0\\
\bottomrule
 \multicolumn{6}{c}{* Data taken from \citet{jelassi2023length}. \tablefootnote{An encoder-only architecture with shared layers.}}  \\
\end{tabular}
\end{center}
\end{table}

%
%

A few interesting patterns emerge. First, since the model generates output tokens in a reversed order, most of the open elements are along the anti-diagonal direction for both tasks. Second, there is a clear division of labor among the heads, which is consistent with the findings in \ref{sec.attn}. More specifically, in \textbf{\texttt{Addition}}, heads $1$, $4$, $7$ attend to the first operand, while the remaining heads attend to the second. In $N \times 1$, most heads attend to the multi-digit number and the multiplication sign while one of the heads, head 4, attends to the single-digit operand. Note that there are vertical lines in heads 1, 3, and 7 as well. Third, the different patterns show that our bias generation process is effective: the anti-diagonal and vertical patterns are learned by searching the corresponding directions.  Note that there is an empty bias consisting of all $0$s in \ref{fig:ABCbiasmultcross} (head 5). This indicates that ABC did not pick up any patterns in that head. 

\textbf{Running Time}. ABC requires a retraining stage. However, with the help of attention bias masks, this stage converges very fast. We observe that the time needed to retrain the model is only 1/100 to 1/10 of the time for training the model to interpolate.

\section{Connections to Other Works}

It turns out that  ABC has close ties to other schemes of manipulating attention. We elaborate on two in the following.

\subsection{ABC as a Generalized RPE}
\label{sec.abc_rpe}

The relative position encoding (RPE) of \citet{shaw-etal-2018-self} has been shown to be a very robust PE and the foundation of many other variants \citep{10.1162/coli_a_00445}. \citet{shaw-etal-2018-self} biases the attention at two places: (1) when computing the dot-product between query and key; and (2) when producing the weighted sum of value vectors. (2) has been shown to be not very useful  \citep{shaw-etal-2018-self}. Let $x_i$ be the embedding vector of the $i$-th token, (1) is implemented as follows: 
\begin{align}
\centering
e_{ij}&=\frac{(x_i\mW^Q)(x_j\mW^K+a_{ij}^{K})^\top}{\sqrt{d_k}}  \nonumber \\
a^K_{ij}&=w_{clip{(j-i, k)}}. \nonumber
\end{align}
where $w$'s are a set of learned vectors and the bias vector $a^K_{ij}$ is selected from the set by a clipped indexing scheme:  $clip(x, k) = \max(-k, \min(k, x))$. That is, tokens more than $k$ units from the current query token will be clipped to $k$. Note that the selection of $w$ vector depends solely on the relative distance between the query token $i$ and the key token $j$. 

It is clear that both RPE and ABC bias the attention matrix. In the case of RPE, this is done by a vector inside the dot-product, whereas ABC achieves this with a scalar bias on the exterior. If we view elements in the bias matrices and which parameter determines each of them, then we can see the remarkable similarities between RPE and ABC. Figure \ref{fig.rpe_vs_abc} shows a comparison between attention bias matrices of RPE and ABC for the case extending along the diagonal.  ABC averages along each of the $k$-diagonals at step 2.1 during its procedure. Thus for query $i$ and key $j$, the bias is $d_{j-i}$. The indexing scheme is exactly the same as that of RPE. And there is an implicit clipping too: for an attention matrix of dimensions $m \times n$ with $m \leq n$,  the set of possible $k$ values for valid $k$-diagonals are $\{-(m-1), -(m-2), \ldots, -1, 0, 1, \ldots, (n-1)\}$, a total of $m+n-1$. When extending to $M \times N$, any elements outside those lines are set to $-\inf$. Effectively, this is an asymmetric clipping function: $clip(j-i, m-1, n-1)$.
%
%
\begin{figure}
\hspace*{1cm}
\begin{subfigure}[t]{0.1\textwidth}
\centering
\begin{tikzpicture}
\matrix[matrix of math nodes,inner sep=1pt,row sep=1em,column sep=.6em, left delimiter={[},right delimiter={]}] (M)
{
  w_0 & w_1 &  w_2   \\
    w_{-1} & w_0 & w_1  \\
     w_{-2} & w_{-1} & w_0  \\
};
\end{tikzpicture}
\end{subfigure}
\hspace*{2cm}
\begin{subfigure}[t]{0.1\textwidth}
\centering
\begin{tikzpicture}
\matrix[matrix of math nodes,inner sep=.4pt,row sep=1em,column sep=.6em, left delimiter={[},right delimiter={]}] (M)
{
  d_0 & d_1 &  d_2   \\
    d_{-1} & d_0 & d_1  \\
     d_{-2} & d_{-1} & d_0  \\
};
\end{tikzpicture}
\end{subfigure}
\caption{Factors determining bias weights in RPE (left) and ABC (right).}
\label{fig.rpe_vs_abc}
\end{figure}

One difference is that RPE learns these parameters during training, whereas in ABC, the biases are calculated from correct interpolation results. By scanning more directions, ABC has the potential to discover more regularities.

\subsection{ABC and LoRA}

Low-Rank Adaptation, or LoRA \citep{hu2021lora} is a prevailing method for fine-tuning LLMs for domain adaptation. LoRA freezes the pre-trained model weights and implements trainable weights as the products of low-rank matrices for each
layer of the Transformer architecture, greatly reducing the number of parameters that need to be trained for downstream tasks.  Interestingly, LoRA also uses additive components to adapt the attention matrices. If LoRA is applied to the attention matrices $\mW^{Q}$ and $\mW^{K}$, the attention weights become
\begin{equation}
\label{eqn.lora}
\displaystyle A_{i,j}={x_i\displaystyle (\mW^{Q}+\Delta \mW^{Q})({\displaystyle \mW^{K}}+\Delta \mW^{K})^\top x_j^\top} 
\end{equation}
where $\Delta \mW^{Q}$ and $\Delta \mW^{K}$ are implemented as the products of two low rank matrices which are obtained via training.

Comparing equations \ref{eqn.abc} and \ref{eqn.lora}, it is clear that: (1) both bias the attention weights; and (2) ABC's bias does not directly depend on current inputs and weight matrices. Instead, it is computed from a selected set of inputs.

\section{Discussion}
\label{sec.diss}

It is important to clarify the scope and limitations of our work. Ours is an initial attempt to study the roles of attention in Transformer's 
inductive learning. Due to \citet{hahn-2020-theoretical} limitation, Transformer cannot learn simple tasks without proper bias. The successful cases obtained in this paper, either via ABS or ABC,  even though only on simple recurrent patterns and task-specific models,  solve a few long-standing difficult or even ``impossible'' tasks (e.g., \textbf{\texttt{Parity}}) and represent a significant step forward.  

In order for ABC to work, the vanilla model must be able to interpolate, achieving near perfect accuracy within the training lengths. Among the tasks we study, ABC does not solve \textbf{\texttt{Parity}} because the vanilla model does not interpolate, even with scratch pad.  More complex tasks, such as multi-digit multiplication, appear to require composition of those simple patterns. Learning such compositions may require scaling up the model dramatically, as the heuristic in the NDR paper \citep{csordas_neural_2021} indicates. 

In its current form, we do not expect ABS or ABC to work directly on more complex tasks such as those LLMs are facing. However, we believe that the insight obtained from this study, that attention biasing is an effective way to enforce inductive bias which is \emph{necessary} for successful inductive learning, points out promising directions. The connections between ABC and RPE/LoRA indicate that ABC could have potential applications in other fields such as NLP as well. 
In this case, a ``soft'' variant of ABC, where ABC's bias is combined with the model's original state, might be more suitable.  ABC can be seen as extracting useful internal representations from the model and utilizing it in future tasks. Similar approaches have been shown to be effective in many NLP tasks such as  \citet{Khandelwal2020Generalization} which, instead of the attention, biases the model's output distribution, and \citet{wang2024greater} which uses a temporary LoRA module \citep{hu2021lora} to embed the internal state of earlier text segments to preserve contextual knowledge and improve the inference of long text. Their successes suggest that there might be opportunities for further explorations.

\bibliography{main}
\bibliographystyle{icml2024}

\newpage
\appendix

\onecolumn

\section{Attention Bias Scaffolding Details}
\label{sec.right_attn}

In this section we provide more details on the importance of attention and our attention bias scaffolding methods. We develop our ideas through a series of initial experimentations, attention weights analysis, and final verification.

Existing approaches to optimizing length generalization of Transformer models have been focusing on two aspects: positional encoding (PE) or/and attention bias (AB). The two concepts are closely related. In fact, we believe they should be treated as two sides of the same coin: All PEs influence the attention, and almost all ABs, with the exception of no PE at all such as \citet{kazemnejad2023impact} and ours, rely on position information to determine the bias. However, the best-performing AB methods' dependency on positional information is indirect: the bias is often determined by the distance between tokens, instead of their positions. Examples include ALiBi \citep{press2022train} and RPE \citep{shaw-etal-2018-self}. In addition, as our ABS and ABC schemes show, AB can work well without any position information. This is consistent with the findings of some previous works. For example, although Transformer's attention mechanism is order-invariant, decoder-only Transformers with causal attention mask are \emph{not} and can model sequences without explicit position information \citep{tsai-etal-2019-transformer}. 

\subsection{Our Thoughts and Findings}

We have an interesting finding in a similar tune. That is, with our mechanism that enables the model to attend to the correct tokens, explicit position encoding is indeed not always necessary, even for achieving perfect generalization. With our architecture, cross-attention allows the model to attend to the correct input while self-attention relays the information from the previous step.  

This leads us to believe that positional encoding or embedding is not the key to achieving good generalization. The right attention is. PE and AB are just means to attain the latter. Since there is no universal PE or AB that generalizes well on all tasks, for the tasks that we study in this work, auxiliary means that target directly at the attention could be used to achieve better generalization. 

\subsection{Initial Experimentation}
\label{sec.results}
            
To develop our ideas, we first train vanilla Transformers with some commonly used length generalization methods, including the original sinusoidal positional encoding, ALiBi, and RoPE, and examine the results. 

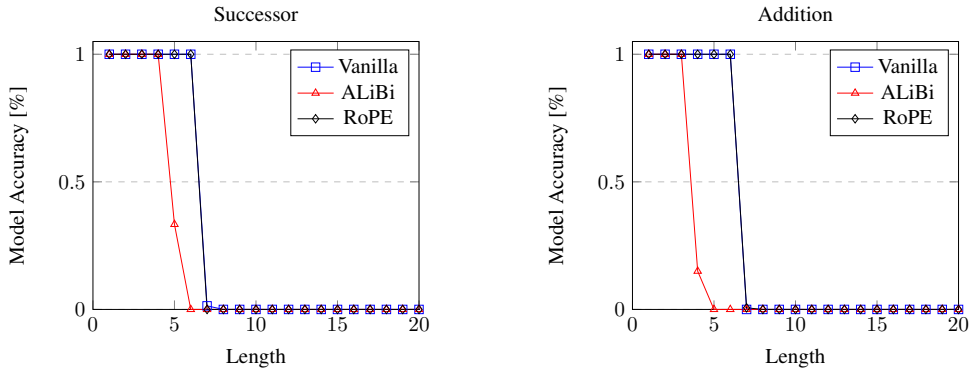
\begin{figure}[H]
\centering
\begin{subfigure}[b]{0.5\textwidth}
\centering
\begin{tikzpicture}[scale=.8]
\begin{axis}[
title={Successor},
    xlabel={Length},
    ylabel={Model Accuracy [\%]},
    xmin=0, xmax=20,
    ymin=0, ymax=1.05,
    legend pos=north east,
    ymajorgrids=true,
    grid style=dashed,
]
\addplot[color=blue,mark=square] table[y= acc_vanilla]{cnt_extrapolation.dat}; \addlegendentry{Vanilla}
\addplot[color=red, mark=triangle] table[y=acc_alibi] {cnt_extrapolation.dat}; \addlegendentry{ALiBi}
\addplot[color=black, mark=diamond] table[y=acc_rope] {cnt_extrapolation.dat}; \addlegendentry{RoPE}

\end{axis}
\end{tikzpicture}
\end{subfigure}%
\hfill
\begin{subfigure}[b]{0.5\textwidth}

\begin{tikzpicture}[scale=.8]
\begin{axis}[
title={Addition},
    xlabel={Length},
    ylabel={Model Accuracy [\%]},
    xmin=0, xmax=20,
    ymin=0, ymax=1.05,
    legend pos=north east,
    ymajorgrids=true,
    grid style=dashed,
]
\addplot[color=blue,mark=square] table[y= acc_vanilla]{add_extrapolation.dat}; \addlegendentry{Vanilla}
\addplot[color=red, mark=triangle] table[y=acc_alibi] {add_extrapolation.dat}; \addlegendentry{ALiBi}
\addplot[color=black, mark=diamond] table[y=acc_rope] {add_extrapolation.dat}; \addlegendentry{RoPE}

\end{axis}
\end{tikzpicture}

\end{subfigure}%
\caption{Extrapolation results for models trained on $L_\mathit{int} \leq 6$ on \textbf{\texttt{Successor}} and \textbf{\texttt{Addition}}. Length is measured in the number of digits of one operand.}
\label{fig.initial}
\end{figure}

Figure \ref{fig.initial} shows the results on \textbf{\texttt{Successor}} and \textbf{\texttt{Addition}}. All models achieve some levels of interpolation but none could extrapolate beyond training length. Among them, RoPE and vanilla Transformer perform almost identically, dropping precipitously to almost 0 accuracy once the length goes beyond 6. Note that the RoPE implementation for \textbf{\texttt{Addition}} must use an embedding size of 512 otherwise it converges very slowly.

We observe similar patterns with other tasks. Table \ref{tab.vannila_int} summarizes  Vanilla Transformer's capabilities for interpolation and extrapolation capabilities on these tasks. We single out the Vanilla model because our ABC scheme works only when the Vanilla model can interpolate.

\begin{table}[h]
\centering
\begin{tabular}{lcc}
\hline
  &\textbf{Interpolation}  &\textbf{Extrapolation} \\ \hline
\textbf{\texttt{Successor}}   & \checkmark & \ding{55}       \\ 
\textbf{\texttt{Addition}}      & \checkmark & \ding{55}     \\ 
\textbf{\texttt{Parity}}      &  \ding{55}  & \ding{55}              \\
\textbf{\texttt{$N \times 1$}}      & \checkmark & \ding{55}                  \\ 
\hline
\end{tabular}
\caption{Vanilla Transformer's interpolation and extrapolation capabilities.}
\label{tab.vannila_int}
\end{table}

\subsection{Attention Analysis}
\label{sec.attn}

\begin{figure}[H]%
\begin{subfigure}{.5\textwidth}
\centering
\includegraphics[height=3.4cm]{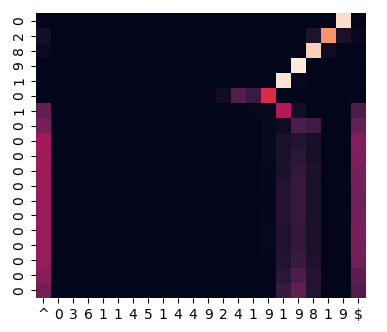} %
\caption{Cross Attention}
\end{subfigure}
\begin{subfigure}{.5\textwidth}
\centering
\includegraphics[height=3.4cm]{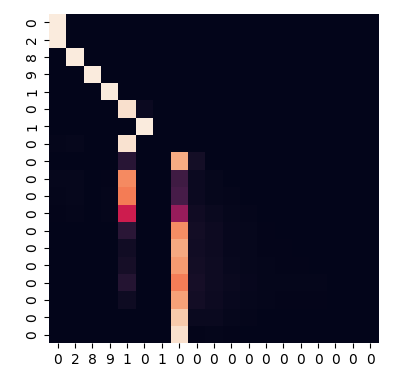} %
\caption{Self Attention}
\end{subfigure}
\caption{Attention heat map on  ``\texttt{03611451449241919819}'' for \textbf{\texttt{Successor}}.}
\label{fig.cntattn}
\end{figure}

To figure out the causes of failure to extrapolate, we extract and analyze the attention weights of the vanilla model on \textbf{\texttt{Successor}} and \textbf{\texttt{Addition}}. Figure \ref{fig.cntattn} gives an example of the attention heat map of one specific head in the last decoder layer during a \textbf{\texttt{Successor}} task. Lighter colors represent higher weights.

For the sequence \texttt{03611451449241919819}, the correct output should be \texttt{03611451449241919820}. Note that we reverse the output digits during training so the model also generates output starting from the lowest digit and working upwards. The model is correct until the hundred-thousands digit. For an input sequence of length $n$, to generate the $i$-th digit for  \textbf{\texttt{Successor}} correctly, the crucial information lies in the $(n-i+1)$-th input token and the $(i-1)$-th output token (for possible carry).\footnote{Note that the tokens are generated in a lowest-digit first order.} This means that the correct attention pattern should light up the ``anti-diagonal'' (the diagonal from top-right to bottom-left) for the cross attention matrix and ``subdiagonal'' (the diagonal directly under the main diagonal) for self-attention. From figure \ref{fig.cntattn} it is clear that the Vanilla Transformer correctly learns the attention pattern up to the hundred-thousands digit and fails beyond that. This correlates perfectly with the extrapolation performance shown in figure \ref{fig.initial}.

\begin{figure}[H]%
\begin{subfigure}{.5\textwidth}
\centering
\includegraphics[height=3cm]{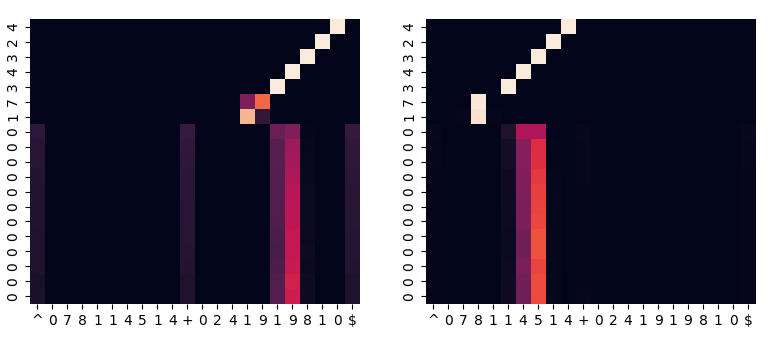} %
\caption{Decoder Cross Attention}
\end{subfigure}
\begin{subfigure}{.5\textwidth}
\centering
\includegraphics[height=3cm]{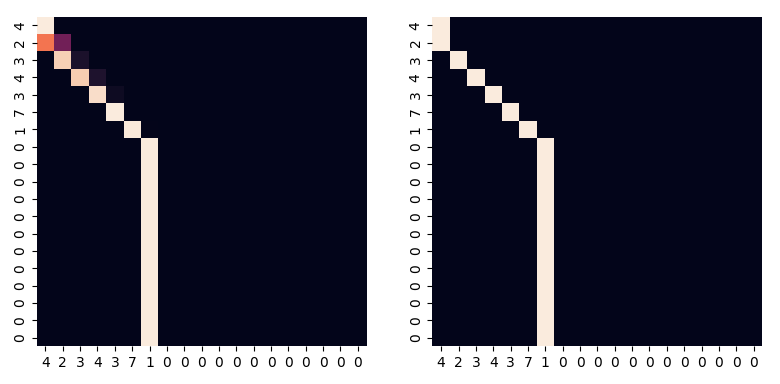} %
\caption{Decoder Self Attention}
\end{subfigure}
\caption{Attention heat map on ``\texttt{078114514+0241919810}'' for \textbf{\texttt{Addition}}.}
\label{fig.attn_add}
\end{figure}

For \textbf{\texttt{Addition}}, we look at individual heads. Figure \ref{fig.attn_add} shows an example of the attention heat maps of two specific heads in the last decoder layer during an addition task. 

In this case we find that there appears to be a sort of differentiation of tasks, where one head looks at the first operand and the other looks at the second. The results are consistent with those found in \textbf{\texttt{Successor}}, that the model does a good job identifying which token to attend to up to the maximum training length. Again this echoes with the extrapolation performance of figure \ref{fig.initial}.


\subsection{Attention Bias Scaffolding}
\label{sec.abs_details}

To future validate our hypothesis, we introduce a number of methods that guide the model to attend to the right places. The ideas are inspired by existing methods for assisting model learning. Those we find effective in arithmetic learning include the following:

\textbf{Input Alignment}

When we humans perform arithmetic computations, input alignment is a common practice that facilitates the process. For example, for multi-digit addition, we write the numbers one below the other, aligning them based on place value. We then add from the rightmost digit, propagating through the left, memorizing carries. Without PE/AB, the original Transformer's attention is order-invariant, and, theoretically, the importance of context does not depend on recency. However, certain input representations result in simplified attention patterns that can be captured by the windowed biasing introduced next. Therefore we interleave the digits from the two operands for binary operations so that digits from each operand that should be attended to together are adjacent. Specifically, for a binary operator $\oplus$ (such as +), and two $n$-digit numbers $a = a_na_{n-1}\ldots a_1$ and $b = b_nb_{n-1}\ldots b_1$ where $a_i$ and $b_i$ are their digits in the proper base representation, the input sequence is transformed as 

\begin{displaymath}
a_na_{n-1} \ldots a_1 \oplus b_nb_{n-1} \ldots b_1 \longrightarrow  \oplus a_nb_na_{n-1}b_{n-1} \ldots a_1b_1
\end{displaymath}

$N \times 1$ is different since the second operand, say $b$, is single-digit. In this case, we just insert $b$ into the right side of each digit of $a$:

\begin{displaymath}
a_na_{n-1} \ldots a_1 \times b \longrightarrow  \times a_nba_{n-1}b \ldots a_1b
\end{displaymath}

 Note that input alignment is only used for ABS, to make the attention pattern simply so subsequent methods could ``scaffold'' the attention to longer inputs more easily. We do \emph{not} need to use it for ABC because ABC could automatically learn the correct patterns. The input to the ABC model is simply the ``natural'' expression (e.g., \texttt{0123+0456} or \texttt{0123*6}).

\textbf{Windowed Attention Biasing}

Biasing towards recency and penalizing attention scores between distant query-key pairs is the basic idea of ABs such as ALiBi \citep{press2022train}. The windowed attention biasing developed by Longformer \citep{beltagy2020longformer} uses a sliding window to control which parts of the attention matrix are ``open''.  We can customize it according to the attention patterns we want to enforce.  

\begin{figure}[H]
\centering
\includegraphics[height=5cm]{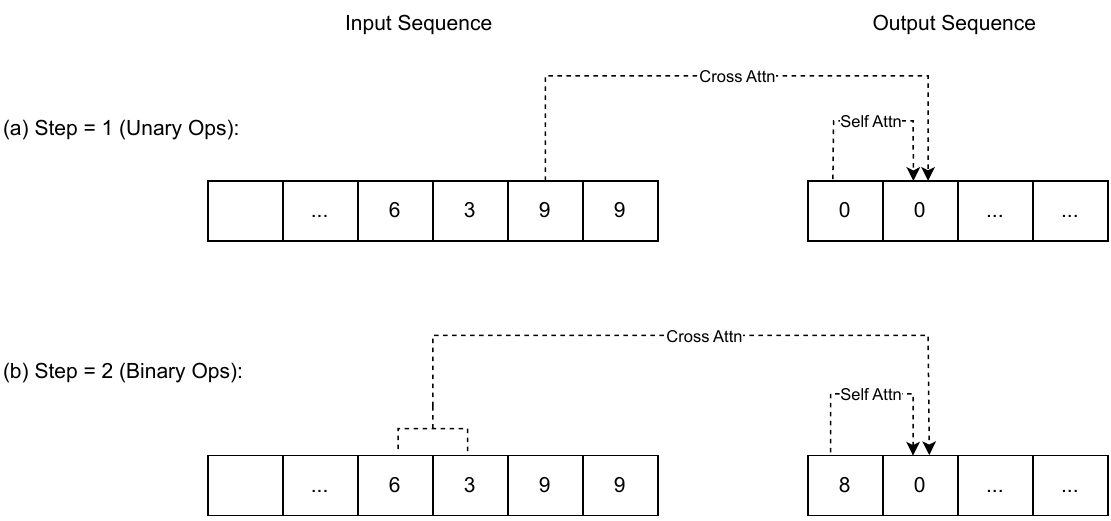}
\caption{Attention patterns for unary and binary operations.}
\label{fig.windowed_attn_pattern}
\end{figure}

Specifically, recall that omitting head indexing, given query, key, and value matrices, the Transformer model \citep{attentionisallyouneed} computes attention scores as:
\begin{displaymath}
Attention(\mQ,\mK,\mV ) = \softmax(\frac{\mQ\mK^T}{\sqrt{d}})\mV
\end{displaymath}
Let $\mA_0 = \frac{\mQ\mK^T}{\sqrt{d}}$ be the original attention weight matrix before softmax, we bias the weights by $\mA = \mA_0 + \mB_w$, where $w$ is a parameter specifying the window width. The basic idea for constructing $\mB_w$ is setting a subset of its elements to 0, and the rest to $-\inf$. This essentially masks out certain elements of $\mA$ to $-\inf$, which, after softmax, results in 0 weights for corresponding tokens,   

The construction of $\mB_w$ depends on the recurrent pattern that encodes the inductive bias about the task \citep{beltagy2020longformer}.  Figure \ref{fig.windowed_attn_pattern} shows the patterns for our tasks. For unary operations, such as successor and parity, generating the current output token depends on the previous output token and one input token at the corresponding position, shown by figure \ref{fig.windowed_attn_pattern} (a). Binary operations, such as addition and $N \times 1$, share the same output token dependency but different input token dependency. In this case, since we align digits from the two operands, as shown in figure \ref{fig.windowed_attn_pattern} (b), the context window spans two consecutive input tokens and also slides two positions at a time.  
For an input length $S$ and output length $L$, the bias for decoder self-attention is 
\begin{displaymath}
B_w = 
 \begin{cases}
	0,& \text{if } i - k = j \text{ for } i, j = 1, \ldots, L, k = 0, \ldots, w \\
	-\inf, & \text{otherwise}
\end{cases}
\end{displaymath}

That is, the elements of the matrix are all set to $-\inf$ except those on the main diagonal and $w$ elements below. Note that, following the traditional practice \citep{attentionisallyouneed} of decoder masking, all elements above the main diagonal are set to $-\inf$ to prevent the decoder from seeing future tokens.

Cross attention bias is similar, with three differences: (1) Since the order of output sequence is reversed, the ``open'' context windows go along the anti-diagonal direction; (2) Since we align the input digits, the window spans, and also steps over, two positions for binary operations; (3) The open context window extends to both left and right $w$ positions. \footnote{Self attention bias only extends to the left.}

\begin{figure}[H]
\centering
\includegraphics[height=8cm]{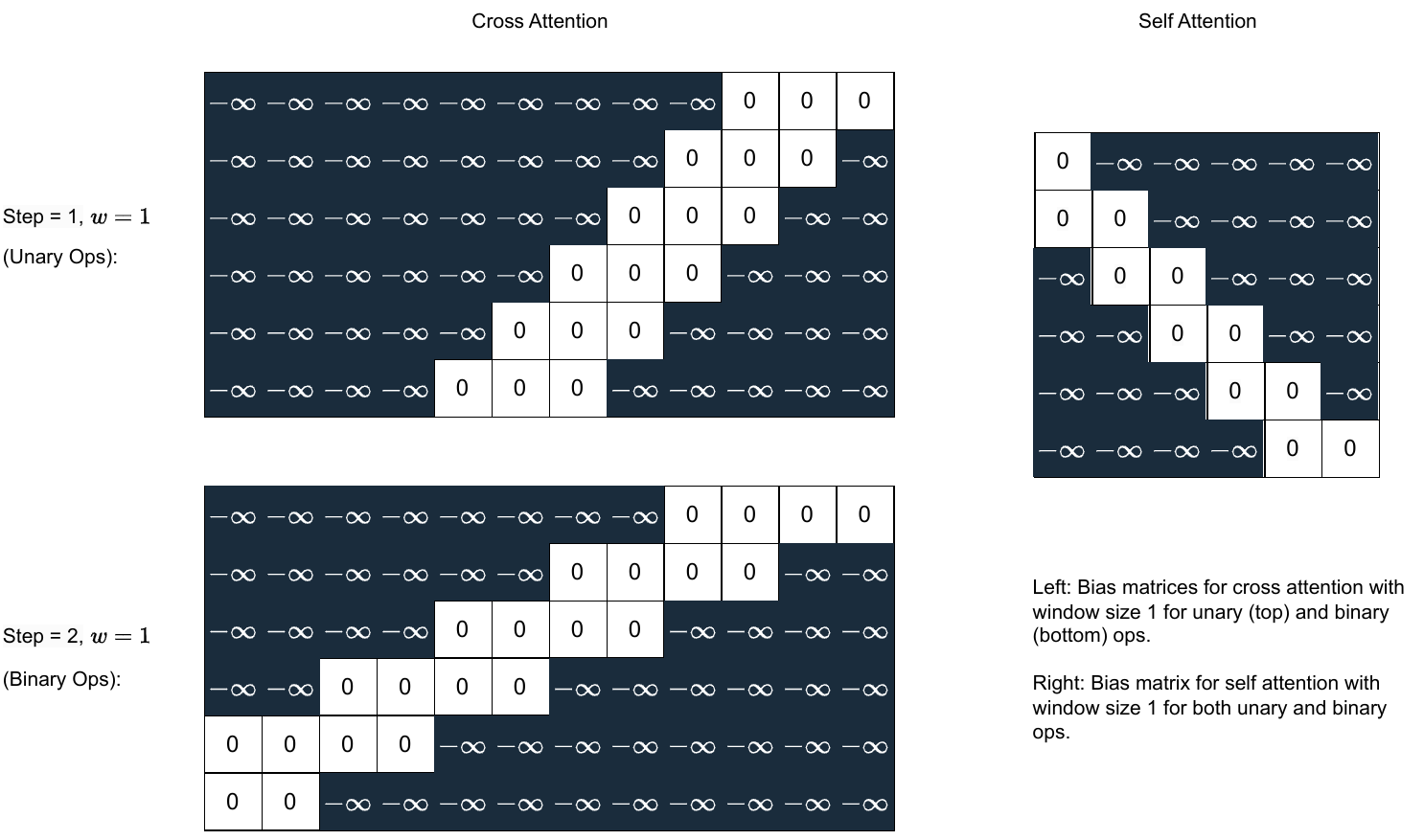}
\caption{Attention bias matrices for unary and binary operations.}
\label{fig.attn_bias}
\end{figure}

Figure \ref{fig.attn_bias} is a visualization for the case of $w = 1$.  

\textbf{Cyclic Position Indexing (CPI)}

Position indexing refers to how we identify each individual position. The simplest way is just to index them $0, 1, \ldots$. Positional embedding mechanisms are then constructed based on this indexing. Very recently, manipulating position indexing has become an effective and trending method for expanding context windows for Transformer-based LLMs. For example, \citet{chen2023extending} and its NTK-aware variant \citep{peng2023yarn} modify RoPE with ``interpolated'' position indices to increase the ``density'' of positions within the pre-trained context window, thus effectively extending its length. 

The motivation for using CPI in our tasks is that large position indices unseen during training may confuse the model. And for arithmetic tasks that admit recurrent generation rules, it is not necessary to identify tokens that are not being currently attended to either. As long as the period is compatible with the context window, it should provide the model with a clear mechanism to differentiate the relevant tokens without diverting its attention.  For arithmetic tasks, our empirical study shows that the model is not sensitive to the value of $T$ as long as it produces an open window whose width is approximately that of the bias context window as shown in figure \ref{fig.windowed_attn_pattern}. We believe it might be of independent interest for other application secenarios.

\subsection{Validation Results}
\label{sec.abs_results}

To evaluate the effectiveness of the above mechanisms, we conduct extensive experiments on each of the arithmetic tasks with the following configurations: 

\begin{enumerate}[label=(\Alph*)]
\item \textbf{Vanilla}: Vanilla Transformer with sinusoidal PE.
\item  \textbf{+ $w = 1$}: (A) + windowed attention biasing with $w = 1$.
\item  \textbf{+ $T = 3$}:  (B) + additional CPI with a period of   $T = 3$. 
\item  \textbf{NoPE + $w = 1$}:  Windowed attention biasing only, without PE at all, with $w = 1$. 
\end{enumerate}

We experimented with a few different $w$ and $T$ values and found that slight variations do not produce very different results thus we report the best-performing configurations above. 

Results are presented in table \ref{tab.abs}. None of the previous works achieves extrapolation on any of the tasks. RPE \citep{jelassi2023length} maintains 90+\% accuracy up to 20 digits but does not go beyond. Vanilla Transformer and RoPE could interpolate, achieving 100\% accuracy for 6-digit inputs, for all the tasks. ALiBi does not even interpolate. Its accuracies drop to near 0s on all tasks beyond 3 or 4 digits (figure \ref{fig.initial}).

On the other hand, our solutions (windowed attention biasing + CPI) achieve complete length generalization on all tasks, maintaining 100\% accuracy up to 60 digits. Unary tasks (\textbf{\texttt{Successor}} and \textbf{\texttt{Parity}}) appear to be not relying on any positional embedding at all once the windowed attention biasing is in place, which is also robust against possible perturbation of any PE.

For binary tasks (\textbf{\texttt{Addition}} and \textbf{\texttt{$N \times 1$}}), on the other hand, there appears to be some bad interaction between the original sinusoidal PE and windowed attention biasing. Both the original sinusoidal PE and $+w = 1$ (sinusoidal PE with windowed bias) configurations only achieve interpolation but not extrapolation. Windowed biasing without any PE at all (NoPE+$w = 1$) results in a slightly imperfect generalization for both binary tasks.

For the \textbf{\texttt{Parity}} task, we list results from the vanilla Transformer attacking it both as a classification problem (outputting 0 or 1), and as a sequence-to-sequence problem (+ scratch pad). Neither works very well. The classification performance is close to random guess while the sequence-to-sequence results are worse. We believe both are attributed to the limitation of \citet{hahn-2020-theoretical}.
Even with a scratch pad, without any attention bias, the generation of each intermediate bit still depends on all the tokens of the input sequence. Furthermore, since obtaining the correct final result depends on all intermediate bits being correct, the task is actually harder due to the compounding-of-errors effect, as the empirical results of table \ref{tab.abs} show.


\begin{table}
\caption{Extrapolation results measured as percent accuracy (\%).  Numbers in bold show the best accuracies achieved for the corresponding input length limit.}
\label{tab.abs}
\begin{center}

\begin{tabular}{c r c c c c c}
\toprule
\multicolumn{2}{c}{\bf}  &\multicolumn{5}{c}{\bf Length (Number of Digits)} \\ \\
\multicolumn{1}{c}{\bf Task} &\multicolumn{1}{c}{\bf Model}  &\multicolumn{1}{c}{\bf 6}  &\multicolumn{1}{c}{\bf 10}   &\multicolumn{1}{c}{\bf 15} &\multicolumn{1}{c}{\bf 20} &\multicolumn{1}{c}{\bf 60} \\
\midrule
    		    &Vanilla  & \bf 100.0 & $0.0$ &$0.0$ &$0.0$ &$0.0$ \\
		    & + $w = 1$ &\bf 100.0 &\bf 100.0 &\bf 100.0 &\bf100.0 & \bf 100.0\\
		     & + $T = 3$ &\bf 100.0 &\bf 100.0 &\bf 100.0 &\bf100.0 & \bf 100.0\\
Successor		     & NoPE + $w = 1$ &\bf 100.0 &\bf 100.0 &\bf 100.0 &\bf100.0 & \bf 100.0\\
  	&ALiBi &$1.3$ &$0.0$ &$0.0$ &$0.0$ &$0.0$\\
           	  &RoPE &\bf 100.0 &$0.0$ &$0.0$ &$0.0$ &$0.0$ \\
\midrule
    		&Vanilla & \bf 100.0 & $0.0$ &$0.0$ &$0.0$ &$0.0$ \\
		& + $w = 1$ &\bf 100.0 & 0.0 & 0.0 &0.0 & 0.0\\
	         & + $T = 3$ &\bf 100.0 &\bf 100.0 &\bf 100.0 &\bf100.0 & \bf 100.0\\
Addition	         & NoPE + $w = 1$ &99.95 &99.81 &99.84 &99.76 & 99.35\\	         
	  	&ALiBi &$0.0$ &$0.0$ &$0.0$ &$0.0$ &$0.0$\\
   	&RoPE &\bf 100.0 &$0$ &$0$ &$0$ &$0$ \\
		&RPE$^*$ &\bf 100.0&$99.9$ &$97.2$ &$21.3$ &N/A \\
\midrule

    		&Transformer$^{\dagger}$ &  &  & $52.00^{\dagger} / 52.60^{\ddagger}  $ &$$ &$$ \\
		 & +  $scratch pad$ & 29.23 &  0.29 & 0.0 &0.0 & 0.0\\
Parity          & + $w = 1$ &\bf 100.0 &\bf 100.0 &\bf 100.0 &\bf100.0 & \bf 100.0\\
		     & + $T = 3$ &\bf 100.0 &\bf 100.0 &\bf 100.0 &\bf100.0 & \bf 100.0\\
		     & NoPE + $w = 1$ &\bf 100.0 &\bf 100.0 &\bf 100.0 &\bf100.0 & \bf 100.0\\
\midrule
   		&Vanilla & \bf 100.0 & $0.0$ &$0.0$ &$0.0$ &$0.0$ \\
                       & + $w = 1$ &\bf 100.0 & 6.0 &0.19 & 0.0 & 0.0\\
$N \times 1$ 		     & + $T = 3$ &\bf 100.0 &\bf 100.0 &\bf 100.0 &\bf100.0 & \bf 100.0\\
		     & NoPE + $w = 1$ & 99.89 &99.63 &99.49 &99.39 & 98.31\\
		     &RoPE &\bf 100.0 &$0$ &$0$ &$0$ &$0$ \\

\bottomrule


\end{tabular}
\end{center}
* Data taken from \citet{jelassi2023length} which is an encoder-only architecture with shared layers. \\
$\dagger$ Data taken from \citet{deletang2023neural} which evaluates five encodings (none, sin/cos, RoPE, ALiBi, and the relative positional encoding from Transformer-XL) and reports the best-performing variant.\\
$\ddagger$ Data taken from \citet{ruoss-etal-2023-randomized} which uses randomized positional encodings to boost length generalization.

\end{table}




\section{Algorithm for ABC}
\label{sec.alg}

\begin{algorithm}[H]
\caption{Attention Bias Calibration (ABC) for non-negative $\Delta$ \protect \footnotemark} \label{alg:cap}
\textbf{Input}:\\
$\mA_{in}$: The attention tensor with dimensions $[H, m, n]$, where $h$ represents the number of heads and $m$, $n$ represents the number of rows and columns in each attention matrix, respectively. \\
$M, N$: The dimensions of the output bias matrix \\
$\sD$: A set of tuples $(1, \Delta)$. It represents the set of all directions we want to search for patterns. \\
\textbf{Output}: $\tilde{\mA}$, a tensor with the dimensions $[H, M, N]$, representing the bias matrix for each head.\\

\begin{algorithmic}
\FOR{$h = 1$ {\bfseries to} $H$}
\FOR{$(1, \Delta) \in\sD$} \STATE \algorithmiccomment{Iterate Directions}
\FOR{$i = 1$ {\bfseries to} $M$}
\FOR{$j = 1$ {\bfseries to} $N$}
\WHILE{$k + i \leq m \: \textbf{and} \: k\Delta + j \leq n,\ k \in \sZ$}
\STATE $\tilde{\mA}_{tmp}[h][(1, \Delta)][i][j]+=\mA_{in}[h][k + i][k\Delta + j]$
\STATE $size +=1$
\ENDWHILE
\STATE $\tilde{\mA}_{tmp}[h][(1, \Delta)][i][j]/=size$ \algorithmiccomment{Average diagonals (if $size \neq 0$)}

\ENDFOR
\ENDFOR

\FOR{$i \gets 1$ to $M$}
\FOR{$j \gets 1$ to $N$}
\STATE $\tilde{\mA}_{tmp}[h][(1, \Delta)][i][j] \gets \tilde{\mA}[h][i][j] -\max(\tilde{\mA})$ \algorithmiccomment{Normalize}
\ENDFOR
\ENDFOR

\FOR{$i \gets 1$ to $M$}
\FOR{$j \gets 1$ to $N$}
\STATE $\tilde{\mA}_{tmp}[h][(1, \Delta)][i][j] \gets dropout(\tilde{\mA}[h][i][j])$ \algorithmiccomment{Dropout}
\ENDFOR
\ENDFOR
\ENDFOR
\FOR{$i \gets 1$ to $M$}
\FOR{$j \gets 1$ to $N$}
\STATE $\tilde{\mA}[h][i][j] \gets \max(\tilde{\mA}_{tmp}[h][(1, \Delta)][i][j], \tilde{\mA}[h][i][j)$ \algorithmiccomment{Merge directions}
\ENDFOR
\ENDFOR
\ENDFOR \\
\STATE return $\tilde{\mA}$
\end{algorithmic}

\end{algorithm}

\footnotetext{The algorithm for negative $\Delta$ is identical except that, before invoking the same procedure, we translate $\mA_{in}$ $N - n + 1$ elements to the right so that the top-right corners of $\mA_{in}$ and $\tilde{\mA}$ align.}


\end{document}